\documentclass[runningheads]{llncs}

\usepackage[T1]{fontenc}
\usepackage{graphicx,verbatim}
\usepackage{algorithm}
\usepackage{algpseudocode}
\usepackage{amsmath}
\usepackage{amssymb}
\usepackage{tabularx}
\newcolumntype{Y}{>{\centering\arraybackslash}X}
\newcolumntype{C}[1]{>{\centering\arraybackslash}p{#1}}
\usepackage{todonotes}

\usepackage{hyperref}
\begin{document}
\title{Image-Conditioned Diffusion Models for Quality Assurance of Organ-at-Risk Segmentations in Radiotherapy}
\titlerunning{Image-Conditioned Diffusion Models for QA of OAR Segmentations}
%
\begin{comment}  %% Removed for anonymized MICCAI 2025 submission
\author{First Author\inst{1}\orcidID{0000-1111-2222-3333} \and
Second Author\inst{2,3}\orcidID{1111-2222-3333-4444} \and
Third Author\inst{3}\orcidID{2222--3333-4444-5555}}
%
\authorrunning{F. Author et al.}
% First names are abbreviated in the running head.
% If there are more than two authors, 'et al.' is used.
%
\institute{Princeton University, Princeton NJ 08544, USA \and
Springer Heidelberg, Tiergartenstr. 17, 69121 Heidelberg, Germany
\email{lncs@springer.com}\\
\url{http://www.springer.com/gp/computer-science/lncs} \and
ABC Institute, Rupert-Karls-University Heidelberg, Heidelberg, Germany\\
\email{\{abc,lncs\}@uni-heidelberg.de}}

\end{comment}

\author{
C Dronne\inst{1,3} \and
C H Clark\inst{2,3,4} \and
X Loizeau\inst{3} \and
E Miles\inst{5} \and
P Hoskin\inst{5,6} \and
J R McClelland\inst{1}
}

\authorrunning{Dronne et al.}

\institute{
UCL Hawkes Institute, Department of Medical Physics and Biomedical Engineering,
University College London, London, United Kingdom
\and
Department of Medical Physics and Biomedical Engineering,
University College London, London, United Kingdom
\and
National Physical Laboratory, Teddington, UK
\and
Department of Radiotherapy Physics,
University College London Hospital, London, UK
\and
National Radiotherapy Trials Quality Assurance Group (RTTQA),
Mount Vernon Hospital, Northwood, UK
\and
Division of Cancer Sciences,
University of Manchester, Manchester, UK
}
\maketitle              

\begin{abstract}
Accurate organ-at-risk segmentation is essential for radiotherapy planning, but reviewing segmentations is time-consuming and subjective. 
We investigate normative modelling for segmentation error detection in head-and-neck CT, comparing a VAE framework with an image-conditioned segmentation diffusion model. 
Models were evaluated on RADCURE brainstem and spinal cord segmentations using simulated boundary and width perturbations. 
Error detection was assessed using the Dice similarity coefficient and the Distance to Agreement (DTA) between the input and reconstructed segmentations. 
While both models detected some simulated errors, regional DTA showed that the diffusion model localised subtle boundary errors more consistently. These results support image-conditioned diffusion reconstruction as a promising framework for localised, anatomy-aware segmentation QA.
Code is available \href{https://github.com/clead6/miccai_workshops26}{here}.

\keywords{error detection  \and diffusion models \and radiotherapy.}

\end{abstract}

\section{Introduction}
Accurate delineation of organs at risk (OARs) is essential for radiotherapy treatment planning and clinical trial quality assurance (QA). Segmentation errors can reduce tumour dose coverage and increase toxicity to healthy tissues. In current clinical practice, OAR segmentations may be produced manually, generated by automated models, or created through hybrid AI-assisted workflows, but all require review before clinical use. In clinical trials, where consistency is critical, this review remains a time-consuming and subjective manual process \cite{tsang_assessment_2019}.

Automated segmentation QA aims to support expert review by identifying cases or regions that may require closer inspection. Unlike uncertainty quantification methods, which are typically tied to a specific segmentation model, post-hoc QA methods should be model-agnostic and applicable to manual, AI-generated, or edited segmentations \cite{van_den_berg_uncertainty_2022}. 

Prior work shows that anatomical image context is essential, as shape-only methods are insufficient for assessing whether a segmentation is appropriate for a specific patient anatomy \cite{chen_automated_2015,hui_quality_2018}. Heuristic or summary metrics likewise fail to capture full spatial information and may miss clinically relevant errors \cite{brooks_knowledge-based_2024,luan_machine_2023}.

Normative modelling provides one approach to segmentation QA by learning the distribution of clinically acceptable image–segmentation pairs. Given a CT image and a segmentation under review, the model reconstructs a segmentation consistent with the learned distribution, and discrepancies between the reconstruction and the submission can serve as a QA signal. Previous studies have demonstrated the feasibility of this framework using Variational Autoencoders (VAEs) \cite{dronne_detection_2026,sandfort_use_2021,wang_deep_2020}.

Denoising Diffusion Probabilistic Models (DDPMs) \cite{ho_denoising_2020} offer an alternative generative framework that has achieved strong performance in image reconstruction and anomaly detection \cite{bercea_evaluating_2025,wyatt_anoddpm_2022}. Their ability to generate sharp, spatially faithful reconstructions may be particularly valuable for segmentation QA, where errors often manifest as subtle local boundary deviations.

In this work, we compare a previously established VAE segmentation QA approach with a new image-conditioned segmentation diffusion model. We evaluate whether using a diffusion model improves the detection and localisation of simulated OAR segmentation errors in a 3D patch-based framework.

\section{Methodology}

\subsection{Proposed framework}
Both the VAE baseline and the proposed diffusion model follow a normative modelling framework in which a given input segmentation is reconstructed by the model using CT image anatomical information. The models are trained on clinically approved image--segmentation pairs and learn a distribution of anatomically plausible segmentations. At inference, a segmentation under review that is consistent with this learned distribution should be reconstructed accurately. In contrast, an atypical or erroneous segmentation is expected to be reconstructed less faithfully or be corrected towards a plausible segmentation. The discrepancy between the input segmentation and its reconstruction can therefore be used as a QA signal.

\begin{figure}[h]
    \centering
    \includegraphics[width=\linewidth]{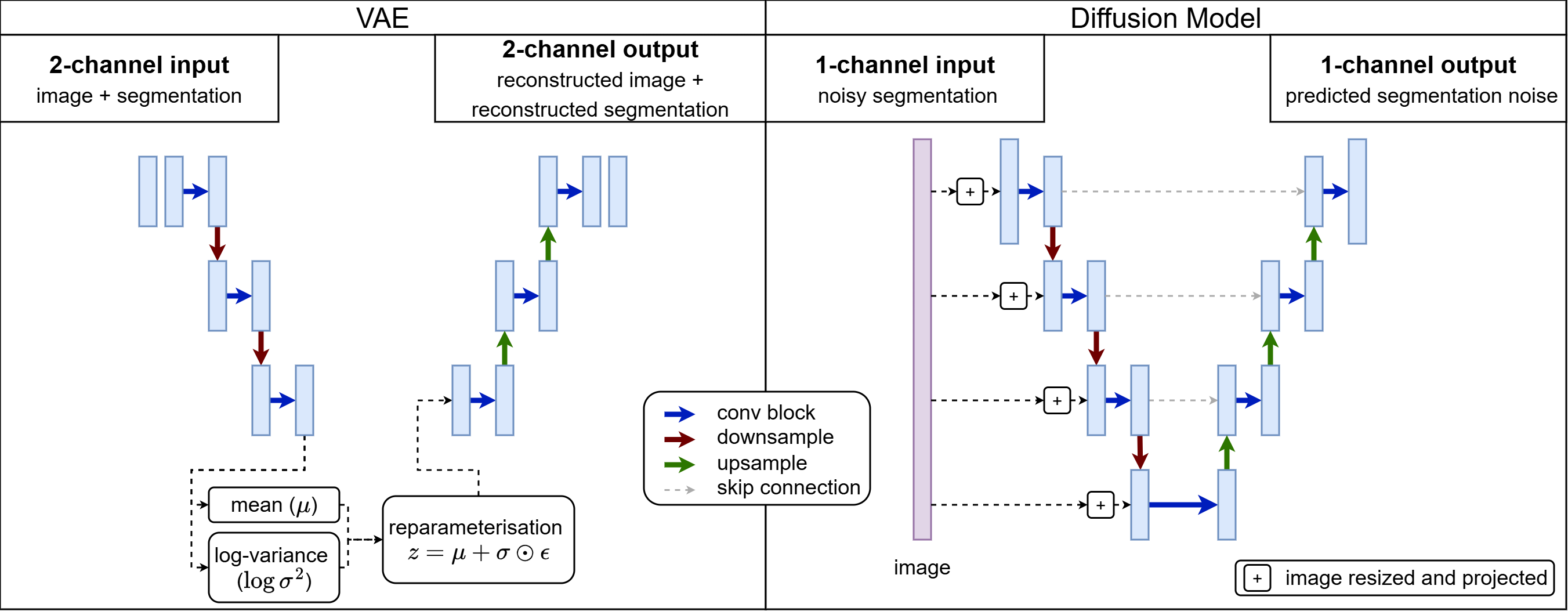}
    \caption{Model architectures compared in this work: image-conditioned segmentation diffusion model and VAE baseline.}
    \label{models}
\end{figure}

This framework is model-agnostic regarding how the input segmentation was generated, making it applicable to manual, AI-generated, and AI-assisted segmentation workflows. The VAE baseline takes the CT image and binary OAR segmentation as multi-channel inputs and reconstructs both channels from a probabilistic latent space \cite{dronne_detection_2026}. In contrast, the proposed diffusion model reconstructs the segmentation after corruption with noise while using the CT image as anatomical conditioning information (Figure~\ref{models}).

\subsection{Diffusion reconstruction framework}
Let $x_0 \in \mathbb{R}^{D \times H \times W}$ denote a CT image and $y_0 \in {0,1}^{D \times H \times W}$ the corresponding input binary segmentation. The proposed model reconstructs the segmentation as $\hat{y}_0$, and segmentation QA is performed by comparing $y_0$ and $\hat{y}_0$. 
Let $T$ denote the total number of diffusion timesteps and let $\{\beta_t\}_{t=1}^T$ denote the diffusion noise schedule, with $\alpha_t = 1-\beta_t$ and $\bar{\alpha}_t=\prod_{s=1}^t \alpha_s$.

The segmentation is corrupted using a Bernoulli diffusion process. At timestep $t$, the corrupted segmentation $y_t$ is sampled voxel-wise as
\begin{equation}
    q(y_t \mid y_0) =
    \mathcal{B}\left(
    y_t ; \bar{\alpha}_t y_0 + \frac{1-\bar{\alpha}_t}{2}
    \right),
\end{equation}
where $\mathcal{B}$ denotes a Bernoulli distribution \cite{chen_berdiff_2023}. Equivalently, this process can be interpreted as randomly flipping segmentation voxels with probability $(1-\bar{\alpha}_t)/2$, so that the mask gradually becomes corrupted by uninformative Bernoulli noise.

The image-conditioned segmentation diffusion model applies diffusion only to the segmentation. The CT image is not corrupted or reconstructed: it remains fixed and is used only to condition segmentation denoising.
The noisy segmentation $y_t$ is passed through a mask-denoising 3D U-Net, while the uncorrupted CT image $x_0$ provides multi-scale anatomical conditioning. 
At each encoder resolution, $x_0$ is resampled to the current segmentation feature map resolution and projected via a lightweight convolutional block to match the width of the corresponding segmentation channel. These projected image features are added to the segmentation features after each encoder block and at the bottleneck. The resulting image-conditioned encoder features are passed to the decoder through the U-Net skip connections and concatenated with decoder features during upsampling. 

The model predicts the Bernoulli flip-noise logits for the segmentation, $g_{\theta}(y_t,x_0,t) = \hat{\epsilon}_y$, where $\epsilon_y$ denotes the voxel-wise Bernoulli flip noise used to corrupt $y_0$ into $y_t$, and $\hat{\epsilon}_y$ denotes the model prediction of this flip noise.

The model is trained with a hybrid diffusion loss derived from the variational lower bound (VLB) on the negative log-likelihood. The loss combines the Bernoulli VLB term with binary cross-entropy (BCE) on the predicted flip noise:
\begin{equation}
L_{\mathrm{cond}} = L_{\mathrm{VLB}}^{y}
+
\mathrm{BCE}(\hat{\epsilon}_y,\epsilon_y).
\end{equation}

At inference, the input segmentation is partially corrupted to timestep $\tau < T$ and denoised to obtain $\hat{y}_0$. Reverse sampling is performed using deterministic Denoising Diffusion Implicit Model (DDIM) sampling \cite{song_denoising_2021}. Reconstructions are obtained at multiple high-corruption timesteps, averaged to produce the final reconstruction, and binarised using a threshold of $0.5$.

\section{Experiments}

\subsection{Dataset and preprocessing}
Experiments were conducted on the RADCURE head-and-neck CT dataset \cite{welch_computed_2024}, focusing on the brainstem and spinal cord as the most frequently segmented organs. CT images and OAR segmentations were converted to NIfTI format, CT intensities were clipped at $+1000$ HU and normalised to $[0,1]$, and all scans were resampled to $2$ mm isotropic voxel spacing. The same preprocessing was used for all models.

Models were trained on $64 \times 64 \times 64$ organ-centred patches. The inferior patch edge was defined using anatomical landmarks: the C2 vertebra for the brainstem and C5 for the spinal cord, segmented with TotalSegmentator \cite{wasserthal_totalsegmentator_2023}. The vertebral centre of mass defined the patch position along the superior--inferior axis, while the in-plane centre was defined with the centre of mass of the corresponding organ segmentation. Brainstem patches contained the full brainstem segmentation. For the spinal cord, where segmentation length varies with scan field of view, patches focused on the superior cord and brainstem--cord interface, a challenging soft-tissue boundary where errors are likely.

Separate models were trained for each organ. Training set size depended on the number of available segmentations: 2,093 cases for the brainstem and 1,993 for the spinal cord. Validation and testing used fixed, organ-independent splits of 472 and 278 cases, respectively.

\subsection{Compared models and implementation}
We compared the published VAE normative modelling approach \cite{dronne_detection_2026} with the proposed image-conditioned segmentation diffusion model. Both models were implemented in PyTorch 2.2.2 and trained on NVIDIA A6000 GPUs. The VAE baseline used the published architecture and training parameters but was trained on the same $64 \times 64 \times 64$ organ-centred patches as the diffusion model.

The diffusion model used a 3D U-Net backbone with timestep embeddings, an initial channel width of 64, channel multipliers $(1,2,4)$, one residual block per resolution level, and dropout of 0.2. It was trained using AdamW \cite{loshchilov_decoupled_2019} with a learning rate of $1 \times 10^{-4}$, a batch size of 8, and early stopping with a patience of 30. The diffusion process used $T=1000$ timesteps with a cosine noise schedule, and DDIM sampling used 50 steps with $\eta=0$.

For the diffusion model, reconstructions were generated at inference timesteps $\tau \in \mathcal{T}_{\mathrm{test}}=\{700,800,900,1000\}$ and averaged to produce a single final reconstruction. Higher $\tau$ values apply stronger corruption, forcing reconstruction to rely more on the learned training distribution. Using multiple high-corruption timesteps improved robustness and reduced dependence on a single simulated error type.

\subsection{QA metrics}
Segmentation QA was based on the discrepancy between the input segmentation and its reconstruction. 
We used voxel-wise Distance to Agreement (DTA) maps to localise regions requiring review. 
DTA was computed over the foreground union of the input and reconstructed segmentations. For voxels present in both segmentations, DTA was set to 0. For voxels present in only one segmentation, DTA was defined as the Euclidean distance, in voxel units, to the nearest voxel where the two segmentations agree. 
Summary DTA values were then computed as the mean DTA over either the full foreground union or the known simulated-error region.
Dice Similarity Score (DSC) between the input and reconstructed segmentations was also reported.

\subsection{Evaluation on simulated segmentation errors}
Segmentation error detection was evaluated by introducing controlled perturbations into the test segmentations, following the published VAE approach \cite{dronne_detection_2026}. Superior ($\uparrow$) and inferior ($\downarrow$) boundary modifications were simulated for the brainstem, while superior ($\uparrow$) boundary modifications and transverse widening ($\leftrightarrow$) were simulated for the spinal cord.

\begin{figure}[H]
    \centering
    \includegraphics[width=\linewidth]{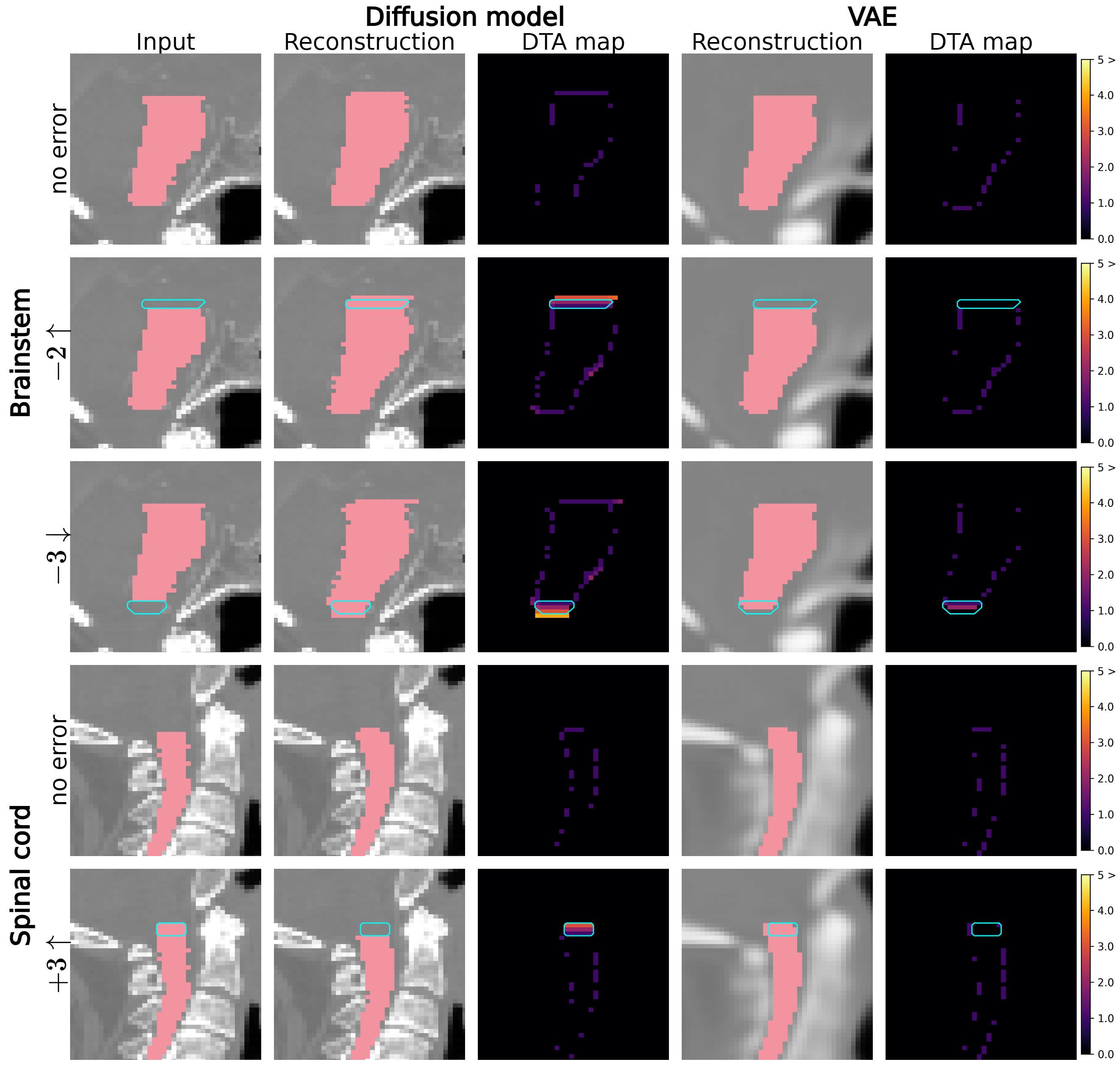}
    \caption{Examples of reconstructions generated by the VAE and diffusion model for simulated brainstem $-2\uparrow$ and $-3\downarrow$ and spinal cord $+3\uparrow$ boundary errors. Top rows show input and reconstructed segmentations overlaid on CT images; bottom rows show corresponding DTA maps. Cyan contours indicate the simulated error region.}
    \label{fig}
\end{figure}

Figure~\ref{fig} shows representative examples of the reconstruction-based QA signal. In the absence of simulated errors, DTA values are low for both models, indicating close agreement between the input and reconstructed segmentations. After perturbations are introduced, higher DTA values appear in the corresponding error regions. This effect is more pronounced for the diffusion model, which produces clear localised DTA responses for all three illustrated perturbations. The VAE shows only a weak response for the $-3\downarrow$ perturbation and limited localisation for the other examples, whereas the diffusion model more clearly highlights the introduced errors.

For each perturbed segmentation, DSC and DTA between the input and reconstructed segmentations were compared with the corresponding original, non-perturbed segmentation. Global metrics, computed over the full segmentation, are reported in Table~\ref{results1}. As Shapiro--Wilk tests indicated non-normality, paired one-sided Wilcoxon signed-rank tests were used to compare each simulated-error condition with the no-error condition.

Global metrics showed the expected overall trend: simulated errors generally decreased DSC and increased DTA as perturbation size increased. However, these full-segmentation summaries provide limited information about whether the discrepancy is localised to the actual error region, which is one of our key requirements for segmentation QA.

We therefore evaluated DTA directly within the simulated-error region and included smaller perturbations. For each perturbation, regional DTA was compared between the original and perturbed segmentations in the same anatomical region using paired one-sided Wilcoxon signed-rank tests. These results are shown in Table~\ref{results2}. An increase in regional DTA indicates that the model not only detects a discrepancy between the input segmentation and its reconstruction, but also localises this discrepancy to the known error region. This regional analysis is particularly relevant for QA, where the goal is not only to flag a potentially erroneous segmentation but also to direct the reviewer to the location requiring inspection.

For subtle brainstem perturbations, the diffusion model showed significant regional DTA increases for the $-2\uparrow$, $-1\uparrow$, $-1\downarrow$, and $+1\downarrow$ perturbations, while the VAE did not significantly detect the $-2\uparrow$ and $-1\uparrow$ perturbations. Neither model showed significant regional DTA increases for the $+1\uparrow$ or $+2\uparrow$ perturbations, suggesting limited sensitivity to these subtle superior extension errors.

\begin{table}[h]
\caption{Global DSC and DTA between input and reconstructed segmentations for original and perturbed brainstem and spinal cord cases. 
The $0$ row indicates no error; $\uparrow$ denotes superior boundary shifts, $\downarrow$ inferior, and $\leftrightarrow$ transverse widening.
Values are mean $\pm$ standard deviation. Bold indicates no significant difference from the no-error condition under a paired one-sided Wilcoxon signed-rank test ($p \geq 0.05$).}
\label{results1}
\centering
\begin{tabularx}{\textwidth}{c | Y Y | Y Y}
\hline
& \multicolumn{2}{c|}{VAE} & \multicolumn{2}{c}{Diffusion model} \\ \hline
& DSC & DTA & DSC & DTA \\ \hline
\multicolumn{5}{c}{Brainstem} \\ \hline
$0$ & $0.94 \pm 0.02$ & $0.15 \pm 0.05$ & $0.92 \pm 0.03$ & $0.32 \pm 0.64$ \\ \hline
$-5\uparrow$ & $0.94 \pm 0.02$ & $0.22 \pm 0.43$ & $0.89 \pm 0.04$ & $0.94 \pm 1.24$ \\
$+5\uparrow$ & $0.92 \pm 0.02$ & $0.24 \pm 0.10$ & $0.90 \pm 0.02$ & $0.39 \pm 0.33$ \\
$-5\downarrow$ & $0.94 \pm 0.02$ & $0.18 \pm 0.12$ & $0.91 \pm 0.03$ & $0.46 \pm 0.66$ \\
$+5\downarrow$ & $0.94 \pm 0.02$ & $0.18 \pm 0.06$ & $0.92 \pm 0.03$ & $0.37 \pm 0.59$ \\ \hline
\multicolumn{5}{c}{Spinal cord} \\ \hline
$0$ & $0.96 \pm 0.01$ & $0.22 \pm 0.06$ & $0.96 \pm 0.01$ & $0.25 \pm 0.09$ \\ \hline
$+5\uparrow$ & $0.96 \pm 0.01$ & $0.31 \pm 0.12$ & $0.95 \pm 0.01$ & $0.40 \pm 0.14$ \\
$-5\uparrow$ & $0.96 \pm 0.01$ & $0.30 \pm 0.10$ & $0.95 \pm 0.01$ & $0.64 \pm 0.24$ \\
$+2\leftrightarrow$ & $0.92 \pm 0.02$ & $0.47 \pm 0.13$ & $0.90 \pm 0.02$ & $0.75 \pm 0.17$ \\ \hline
\end{tabularx}
\end{table}

For the spinal cord, the diffusion model showed significant regional DTA increases for the $+3\uparrow$, $-1\uparrow$, and $+2\leftrightarrow$ perturbations. In contrast, the VAE did not significantly detect the transverse widening perturbation or the $+3\uparrow$ perturbation. These results suggest that the diffusion model provides the most consistent detection and localisation of the simulated errors.

\begin{table}[H]
\caption{Regional DTA within the simulated-error region, compared between original and perturbed segmentations. $\uparrow$ denotes superior boundary shifts, $\downarrow$ inferior, and $\leftrightarrow$ transverse widening.
Values are mean $\pm$ standard deviation. Bold values indicate no significant DTA increase for the perturbed segmentation under a paired one-sided Wilcoxon signed-rank test ($p \geq 0.05$).}
\label{results2}
\centering
\begin{tabularx}{\textwidth}{c | Y Y | c | Y Y | c}
\hline
& \multicolumn{3}{c|}{VAE}  & \multicolumn{3}{c}{Diffusion model} \\ \hline
& Original & Error & p-value & Original & Error & p-value \\ \hline
\multicolumn{7}{c}{Brainstem} \\ \hline
$-2\uparrow$ & $0.11 \pm 0.12$ & $0.17 \pm 0.39$ & $\mathbf{1.00}$ & $0.08 \pm 0.09$ & $1.11 \pm 0.52$ & $<0.001$ \\
$-1\uparrow$ & $0.16 \pm 0.23$ & $0.08 \pm 0.27$ & $\mathbf{1.00}$ & $0.10 \pm 0.16$ & $0.59 \pm 0.50$ & $<0.001$ \\
$+1\uparrow$ & $0.64 \pm 0.10$ & $0.44 \pm 0.20$ & $\mathbf{1.00}$ & $0.65 \pm 0.13$ & $0.41 \pm 0.20$ & $\mathbf{1.00}$ \\
$+2\uparrow$ & $0.57 \pm 0.13$ & $0.41 \pm 0.20$ & $\mathbf{1.00}$ & $0.64 \pm 0.21$ & $0.41 \pm 0.24$ & $\mathbf{1.00}$ \\
$-1\downarrow$ & $0.39 \pm 0.42$ & $0.64 \pm 0.48$ & $<0.001$ & $0.39 \pm 0.69$ & $0.60 \pm 0.50$ & $<0.001$ \\
$+1\downarrow$ & $0.40 \pm 0.49$ & $0.62 \pm 0.53$ & $<0.001$ & $0.26 \pm 0.44$ & $0.68 \pm 0.93$ & $<0.001$ \\ \hline
\multicolumn{7}{c}{Spinal cord} \\ \hline
$+3\uparrow$ & $0.57 \pm 0.54$ & $0.64 \pm 0.64$ & $\mathbf{0.63}$ & $0.60 \pm 0.70$ & $0.93 \pm 0.83$ & $<0.001$ \\
$+2\uparrow$ & $0.57 \pm 0.53$ & $0.54 \pm 0.58$ & $\mathbf{1.00}$ & $0.56 \pm 0.62$ & $0.64 \pm 0.68$ & $\mathbf{0.19}$ \\
$+1\uparrow$ & $0.54 \pm 0.50$ & $0.49 \pm 0.57$ & $\mathbf{0.85}$ & $0.47 \pm 0.50$ & $0.42 \pm 0.56$ & $\mathbf{0.37}$ \\
$-1\uparrow$ & $0.32 \pm 0.39$ & $0.75 \pm 0.44$ & $<0.001$ & $0.20 \pm 0.33$ & $0.76 \pm 0.44$ & $<0.001$ \\
$+2\leftrightarrow$ & $1.01 \pm 0.02$ & $0.68 \pm 0.20$ & $\mathbf{1.00}$ & $1.02 \pm 0.03$ & $1.09 \pm 0.26$ & $<0.001$ \\ \hline
\end{tabularx}
\end{table}

\section{Discussion}
This work demonstrates the potential of image-conditioned diffusion models for QA of OAR segmentations. Compared with the VAE baseline, the diffusion model more consistently detected and localised simulated boundary errors, particularly when DTA was evaluated within the affected region.

The stronger performance of the diffusion model may partly reflect how anatomical information is preserved during reconstruction. The VAE reconstructs the CT image and OAR segmentation through a shared latent representation. However, this compression can produce blurred CT reconstructions, reducing the anatomical detail available to guide segmentation reconstruction. Consequently, the VAE may rely more on generic organ-shape information, limiting its sensitivity to subtle segmentation errors.

A joint diffusion strategy, in which both the CT image and segmentation are corrupted, also presents a related limitation. Effective segmentation QA requires strong corruption of the input segmentation before reconstruction, but applying high noise levels to the CT image degrades the anatomical detail needed for boundary assessment. The image-conditioned diffusion model avoids this by corrupting only the segmentation while keeping the CT image uncorrupted as conditioning information, preserving patient-specific context during denoising and producing DTA maps that more clearly highlight error regions.

An additional consideration is the large segmentation variability in the RADCURE dataset, as discussed in \cite{dronne_detection_2026}. Many segmentations were produced before clear delineation guidelines were available and therefore vary substantially between cases, particularly at soft-tissue boundaries \cite{brouwer_ct-based_2015}. As a result, small simulated perturbations may make a segmentation more consistent with the average anatomy learned from the training data, rather than causing a clear error. This may explain why some subtle slice-addition or slice-removal perturbations did not produce statistically significant regional DTA increases. More broadly, evaluation using simulated perturbations may not capture the full variability of real clinical segmentation errors. Future work will evaluate the method on real clinical segmentation errors and additional OARs.

The proposed approach is intended to support, rather than replace, manual clinical review by directing attention to regions with high reconstruction disagreement, which may indicate possible segmentation errors. This could make OAR segmentation review more focused, consistent, and efficient. 
A practical limitation of the diffusion model is increased inference time: the VAE required 0.5s per image per organ, whereas each diffusion reconstruction required 4.6s and was repeated across four timesteps for aggregation.
However, this runtime is likely acceptable for offline QA workflows, where cases can be processed before clinician review. 
Overall, image-conditioned diffusion reconstruction provides a promising framework for localised, anatomy-aware segmentation QA.

\begin{credits}
\subsubsection{\ackname} This work is supported by the EPSRC-funded UCL Centre for Doctoral Training in Intelligent, Integrated Imaging in Healthcare (i4health) [EP/S021930/1]; the National Physical Laboratory; and the Mount Vernon Marie Curie Research Wing Trust. The National Radiotherapy Trials Quality Assurance (RTTQA) Group is funded by the National Institute for Health and Care Research (NIHR).

\end{credits}

\clearpage
\bibliographystyle{splncs04}
\bibliography{bib}

\end{document}